\documentclass[nohyperref]{article}

\usepackage{microtype}
\usepackage{graphicx}
\usepackage{subfigure}
\usepackage{booktabs} 

\usepackage{hyperref}

\usepackage[accepted]{icml2022}

\usepackage{amsmath}
\usepackage{amssymb}
\usepackage{mathtools}
\usepackage{amsthm}

\usepackage[capitalize,noabbrev]{cleveref}

\theoremstyle{plain}

\theoremstyle{definition}

\theoremstyle{remark}

\usepackage[textsize=tiny]{todonotes}

\usepackage{tikz}
\usepackage{graphicx}
\usepackage{textcomp}
\usepackage{xcolor}      
\usepackage{soul}
\newcommand*\rotate{\rotatebox{-90}}
\def \u {\mathbf u}
\def \v {\mathbf v}
\def \y {\mathbf y}
\def \x {\mathbf x}
\def \blambda {\boldsymbol \lambda}
\def \t {^ \top}

\icmltitlerunning{ICML 2022 HAET Workshop}

\begin{document}

\twocolumn[
\icmltitle{Rethinking Pareto Frontier for Performance Evaluation of Deep Neural Networks}



\icmlsetsymbol{equal}{*}

\begin{icmlauthorlist}
\icmlauthor{Vahid Partovi Nia}{comp}
\icmlauthor{Alireza Ghaffari}{comp,sch}
\icmlauthor{Mahdi Zolnouri}{comp}
\icmlauthor{Yvon Savaria}{sch}
\end{icmlauthorlist}

\icmlaffiliation{comp}{Huawei Noah's Ark Lab, Montral, Canada}
\icmlaffiliation{sch}{Department of electrical engineering, Polytechnique Montreal, Montreal, Canada}

\icmlcorrespondingauthor{Vahid Partovi Nia}{vahid.partovinia@huawei.com }

\icmlkeywords{Machine Learning, ICML}

\vskip 0.3in
]



\printAffiliationsAndNotice{}  

\begin{abstract}
Performance optimization of deep learning models is conducted either manually or through automatic architecture search, or a combination of both. On the other hand, their performance strongly depends on the target hardware and how successfully the models were trained. 
We propose to use a \emph{multi-dimensional} Pareto frontier to re-define the efficiency measure of candidate deep learning models, where several variables such as training cost, inference latency, and accuracy play a relative role in defining a dominant model. Furthermore,  a random version of the multi-dimensional  Pareto frontier is introduced to mitigate the uncertainty of accuracy, latency, and throughput of deep learning models in different experimental setups. These two complementary methods can be combined to perform objective benchmarking of deep learning models. Our proposed method is applied to a wide range of deep image classification models trained on ImageNet data. 
Our method combines competing variables with stochastic nature in a single relative efficiency measure. This allows ranking deep learning models that run efficiently on different hardware, and combining inference efficiency with training efficiency objectively.
\end{abstract}

\section{Introduction}

Deep learning  has revolutionized the artificial intelligence domain by surpassing human performance in a variety of computer vision tasks. 
Image classification models \cite{krizhevsky2012imagenet}  are the center of attention in computer vision, mostly because they  are used as the backbone image representation of other downstream perception tasks such as object detection, and image segmentation. Later, deep learning models  extended to speech \cite{oord2016wavenet} and natural language processing \cite{vaswani2017attention}.

Since deploying deep learning models are compute intensive, researchers actively try to automate the process of finding the best solution for each specific context. Automating the process of finding the most effective solution to a deep learning tasks is associated with three main issues: (i) availability of data required to train the model (ii) time and computational resources required for training (iii) time and computational resources required for inference. A recent trend in neural network optimization is to focus on inference only (e.g. quantization techniques), while data and training aspects are often ignored. Moreover, even when focusing on inference, various measures of performance versus complexity can be defined. For instance, top-1 or top-5 accuracy are often treated as measures of accuracy. Moreover, inference latency, and throughput measured in frame per second (FPS) are metrics that strongly depend on the target hardware. 

Designing low complexity models is often performed manually. This is costly for large corporation that deal with different hardware constraints and intend to automate a variety of tasks using a single deep learning model. Several solutions have been proposed to automate the design process using reinforcement learning \cite{zoph2018learning}, evolutionary methods \cite{shafiee2018deep}, or grid search \cite{tan2019efficientnet}. A combination of manual design with automatic design was also explored recently \cite{wong2019yolo}. Interestingly, hardware designers also started to use similar methods to optimize ML hardware targets. 
For instance, in \cite{chen2020hardware}, the authors used reinforcement learning for resource allocation in deep learning accelerators. In \cite{ghaffari2021efficient}, a hybrid  evolutionary method is used to optimize hardware design. Furthermore, Different designs and frameworks are developed to run neural networks efficiently on CPU \cite{courville2019deep}.

Whether designing a low complexity deep learning model or a high-performance hardware, finding the best solution is costly, time-consuming, and in some cases impractical, as multiple complexity parameters, and architectures must be taken into account. 
Moreover, various constraints and objectives might be defined. For example, in one task we may sacrifice the accuracy to achieve lower power consumption, while the same trade-off may not be valid in other scenarios.

Anther important issue is performance variability of ML systems. Thus, we aim to help solving multi-objective and multi-dimensional design problems with systems implementing algorithms with variable performance.

Multi-objective optimization and Pareto-optimal solutions are known among the  hardware design community. Researchers are often interested to optimize the hardware for better throughput and less power consumption as well as chip area utilization. For example, in \cite{santoro2018design}, the authors used Pareto optimality to adapt the processing throughput of machine learning accelerators to the workload requirements. In \cite{ghaffari2021efficient,gautier2016spector}, Pareto optimal design is used to propose optimal solutions in design space exploration of OpenCL kernels on FPGA targets. Furthermore, in \cite{mehrabi2020bayesian}, Pareto optimal designs are used in the synthesis of accelerators. In contrast to these references, this work proposes a stochastic Pareto frontier to cope with the randomness of deep learning performance.

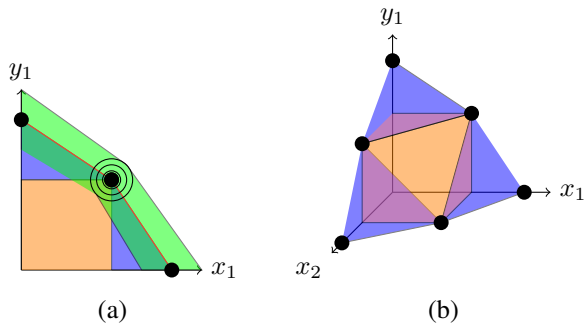
\begin{figure}[!b]
    \begin{tikzpicture}[scale=0.40]
    \draw [->] (0,0) -- (6,0,0) node [right] {$x_1$};
    \draw [->] (0,0) -- (0,6,0) node [above] {$y_1$};
    \draw[fill=blue,opacity=.5] (0,5)--(3,3)--(0,3) node [above right]{$~$};
    \draw[fill=blue,opacity=.5] (3,3)--(5,0)--(3,0) node [above right]{$~$};
    \draw[fill=orange,opacity=.5] (0,0)--(0,3)--(3,3)--(3,0) node [above right]{$~$};
    \draw[fill=green,opacity=.5] (0,4)--(0,6)--(3.5,3.5)--(6,0)--(4,0)--(2.5,2.5) node [above right]{$~$};
    
    \draw[color=red!80] (0,5) -- (3,3) -- (5,0);

    \path (0,5) node[circle, fill, inner sep=2]{};
    \path (3,3) node[circle, fill, inner sep=2]{};
    \path (5,0) node[circle, fill, inner sep=2]{};
    \path (3,3) node[circle, fill, inner sep=2]{};
    
    \draw (3,3) circle (0.7);
    \draw (3,3) circle (0.5);
    \draw (3,3) circle (0.3);
\end{tikzpicture}
~~~~
\begin{tikzpicture}[scale=0.35]
    \draw [->] (0,0) -- (6,0,0) node [right] {$x_1$};
    \draw [->] (0,0) -- (0,6,0) node [above] {$y_1$};
    \draw [->] (0,0) -- (0,0,6) node [below left] {$x_2$};
    \draw[fill=blue,opacity=.5] (3,3,0)--(3,0,3)--(5,0,0) node [above right]{$~$};
    \draw[fill=blue,opacity=.5] (0,3,3)--(3,0,3)--(0,0,5) node [below right]{$~$};
    \draw[fill=blue,opacity=.5] (0,3,3)--(3,3,0)--(0,5,0) node[above right]{$~$};
    \draw[fill=orange,opacity=.5] (0,3,3)--(3,3,0)--(3,0,3) node[above right]{$~$};
    \draw[fill=red!50,opacity=.5] (0,3,3)--(3,3,0)--(0,3,0) node [above left]{$~$};
    \draw[fill=red!50,opacity=.5] (3,0,3)--(0,0,3)--(0,3,3) node [above left]{$~$};
    \draw[fill=red!50,opacity=.5] (3,0,3)--(3,0,0)--(3,3,0) node [above right]{$~$};

    \path (5,0,0) node[circle, fill, inner sep=2]{};
    \path (0,5,0) node[circle, fill, inner sep=2]{};
    \path (0,0,5) node[circle, fill, inner sep=2]{};
    \path (0,3,3) node[circle, fill, inner sep=2]{};
    \path (3,3,0) node[circle, fill, inner sep=2]{};
    \path (3,0,3) node[circle, fill, inner sep=2]{};
\end{tikzpicture}

\hspace{1.3cm}(a)\hspace{4cm}(b)

    \caption{(a) A two-dimensional Pareto frontier with random variations of a point, leading to a random Pareto frontier highlighted in green. (b) A three-dimensional Pareto variant obtained by adding another input to (a).  }
    \label{fig:pareto}
\end{figure}

To summarize, this paper proposes:
\begin{itemize}
    \item A mean to include hardware constraints in the training of deep learning models.
    \item A new tool, the \textit{stochastic multi-dimensional Pareto frontier} that allows mitigating the performance randomness of deep learning models  (Sec.~\ref{sec:pareto}).
    \item To re-define the  \textit{relative efficiency measure} in order to rank deep learning models.(Sec.~\ref{sec:efficiency}).
    \item To define the concept of \textit{stochastic dominance} in order to find the best solution for a given task. (Sec.~\ref{sec:ranking})
\end{itemize}

\begin{figure*}[!t]
    \centering
        \includegraphics[width=0.5\textwidth]{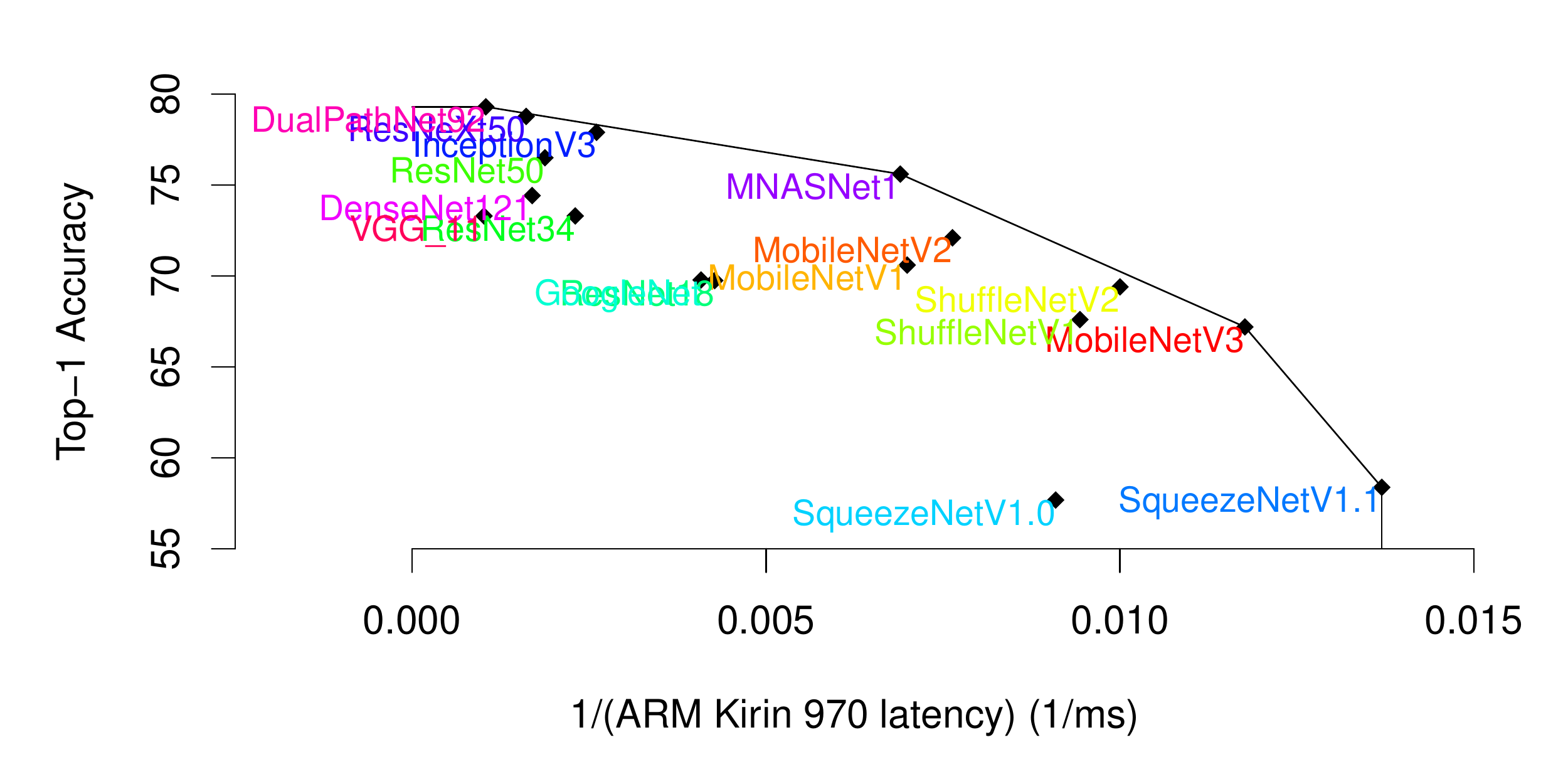}~~
        \includegraphics[width=0.5\textwidth]{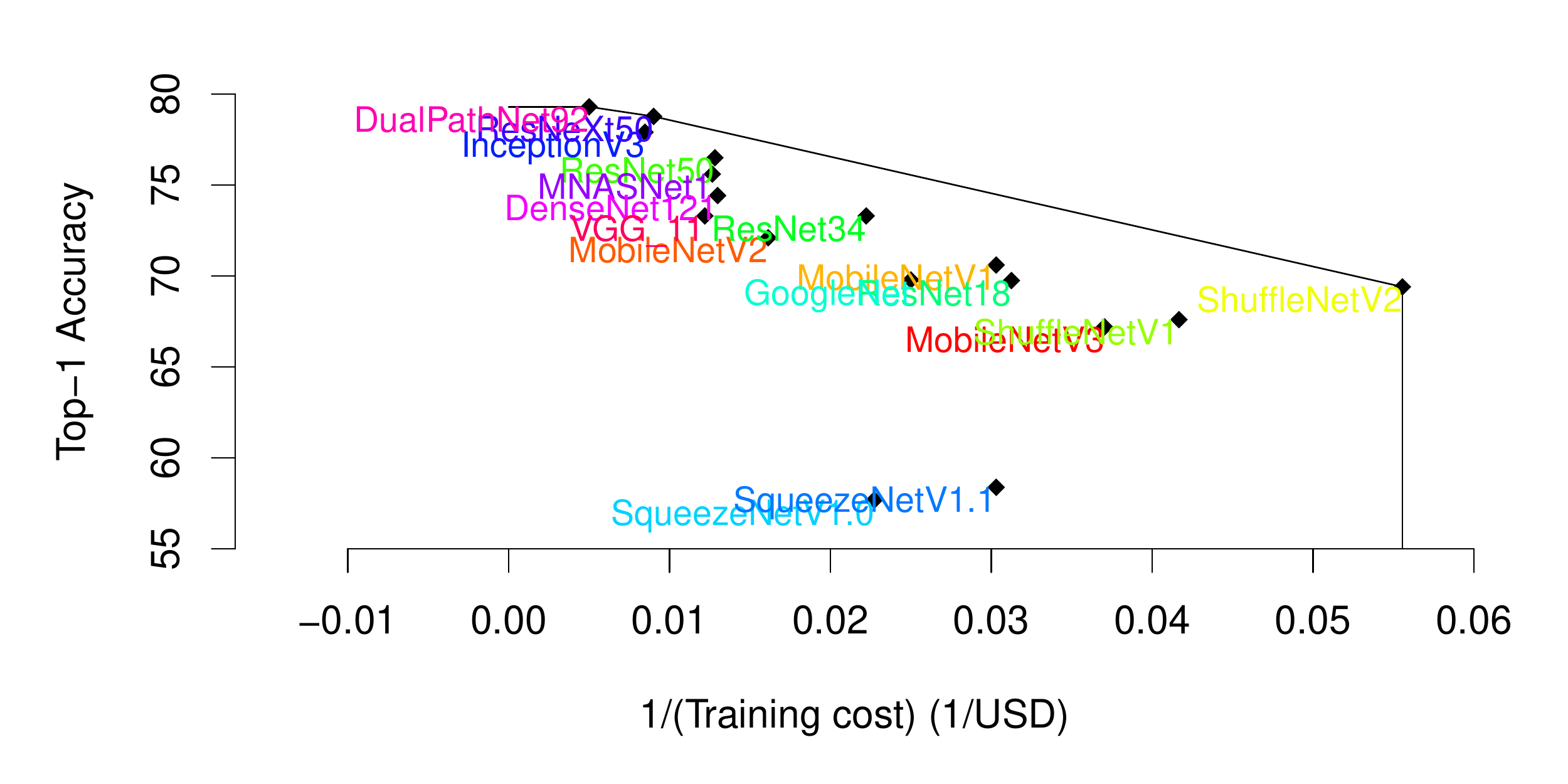}\\
        \small{(a)}\hspace{10cm}\small{(b)}\\

    \caption{ (a) Univariate Pareto frontier for top-1 accuracy versus CPU latency in milli-seconds, (b) Univariate Pareto frontier for top-1 accuracy versus training cost in dollars.
    }
    \label{fig:unipareto}
\end{figure*}

\section{Proposed Methodology}
\subsection{Stochastic Multi-dimensional Pareto Frontier}\label{sec:pareto}
Designing complex systems is often associated with multiple objectives that compete with each other. 
For instance, in the design of a CPU, researchers normally want to minimize the power consumption while maximizing the throughput.
However, naturally, having a faster CPU corresponds to more power consumption. An important question here is: for a \textit{given} power budget, what is the best configuration that maximizes the throughput?

Hence, in multi-objective design problems, the notion of \textit{dominance} comes handy. A solution is dominant if none of the objective functions can improve their scores unless some other objective score decreases. In Figure~\ref{fig:pareto}(a), solutions that are positioned on the red curve are dominant over all the solutions that fall under the red curve. For convenience, we can define the red curve as the Pareto frontier which consists of a set of all dominant solutions. A Pareto frontier shows the interaction between the objectives of the problem. Although in Figure~\ref{fig:pareto}, the Pareto frontier is a curve, Figure~\ref{fig:pareto}(a), it is obvious that when we have three objectives, the curve becomes a ``surface'', Figure~\ref{fig:pareto}(b). 

However, there is a fundamental flaw in using a deterministic Pareto frontier in deep learning. The objective scores related to efficiency measures often include uncertainties. Therefore, the Pareto frontier changes to a \textit{stochastic Pareto frontier}, illustrated in Figure~\ref{fig:pareto}(a) as a green shaded band. A readily available example where this is observed comes from the training of a deep learning model. The training is based on stochastic gradient descent, so it is unlikely to achieve the same top-1 and top-5 accuracy if the training is repeated. The same phenomenon is observable in hardware implementations. For instance, if a compute-intensive algorithm is executed on a CPU or a GPU, it is unlikely to achieve the same latency scores in all execution runs due to operating system interrupts, and the side effects of processing load on the device. 

Based on stochastic Pareto frontiers, some models are significantly better, while others may swap their ranking due to random variations. When significant performance variability are observed, it is important to find models that dominate the others statistically. To that end, we propose to use a Monte Carlo method that samples data from a parametric distribution family, such as a Gaussian distribution, to compensate for the randomness of some observed performance metrics \cite{davtalab2019stochastic}. For example, (i) we can assume a distribution (e.g. Gaussian distribution) for the latency score. Then, we sample the latency of the CPU in different workloads to estimate the distribution parameters such as its mean and variance and (ii) use this Gaussian distribution to estimate the random variation of the relative efficiency. Here, we only focus on Gaussian distribution and upon estimating mean $\mu$ and variance $\sigma^2$ of each experiment, we reconstruct the efficiency distribution using Algorithm~\ref{alg:bootstrap}.
Also after performing  the parametric bootstrap (Algorithm~\ref{alg:bootstrap}, Step 1), there are $B$ samples available for the relative efficiency of setup $i$ that can be used to estimate its random variation. This method is known as \emph{parametric bootstrap} in the statistics literature and is used to estimate complex stochastic phenomena numerically \cite{davison1997bootstrap}. 

\begin{algorithm}[!b]
   \caption{Parametric bootstrap}
   \label{alg:bootstrap}
\begin{algorithmic}
  \FOR{$i = 1, 2~...~,n$}
  { 
    \item \textbf{Step 1)} Measure $k$ times and estimate distribution parameters $(\hat\mu_i, \hat\sigma_i)$ for each experiment setup

  \FOR{$b = 1, 2~...~, B$}
  { 
    \item \textbf{Step 2)} Sample setup $i\in{1,\ldots,n}$ from Gaussian distribution  $\sim N(\hat\mu_i,\hat\sigma_i)$\\
    \item \textbf{Step 3)} Compute relative efficiency $\hat\theta_b$ on this sampled data.
  }\ENDFOR
  }\ENDFOR
\end{algorithmic}
\end{algorithm}

\subsection{Relative Efficiency Evaluation}\label{sec:efficiency}
 Traditionally relative efficiency $\theta_i$ is defined as output $y$, (the higher the better) divided by input $x$ (the lower the better), but normalized to have a unit maximum for objective scores. In other words $\theta_i = {{y_i / x_i} \over \max \left(y_i / x_i\right)}$ assuming both $y_i$ and $x_i$ are univariate positive objective scores. In the presence of multiple inputs $x$ or multiple outputs $y$  a weighted average of these objectives is defined by an expert, to transform them into univariate variables $y_i=\u\t \y_i, x_i = \v\t \x_i$. However, in this approach, $\u$ and $\v$ are subjective, and \cite{banker1984some} proposed to solve the following linear optimization problem as the generalization of relative efficiency in the multi-dimensional case.  

 Suppose $\y_i$ and $\x_i$ are the vectors of outputs and inputs (i.e. objective scores) for experiments  $i$:
 \begin{equation}
    \begin{gathered}
    \theta_i = \min_{\blambda_1 ,\blambda_2, \theta}~ \theta, \\ \mathrm{s.t.~} \blambda_1\t \x_l\leq \theta x_{il}, \quad \blambda_2\t \y_j \geq y_{ij}, \quad \blambda_1, \blambda_2 \geq 0,  \\(\blambda_1, \blambda_2)\t\mathbf 1 = 1,
    \label{eq:vrs_dual}
    \end{gathered}
\end{equation}
in which, $i \in \{1,\ldots,n\}$ is the experiments index and  $j \in \{1,~\ldots~,J\}$ is the dimension of $\y_j$ and $l \in \{1,~\ldots~,L\}$ is the dimension of $\x_i$. The affine Pareto front is obtained by removing the convexity constraint $(\blambda_1, \blambda_2)\t\mathbf 1 = 1$. Note that the univariate affine optimization with $J=1, L=1$ coincides with the classical relative efficiency $\theta_i = {{y_i / x_i} \over \max \left(y_i / x_i\right)}$.
We consider a univariate output score $J=1$, but multiple inputs score.

\section{Ranking Deep Learning Models} \label{sec:ranking}

\begin{table}[!b]
 \caption{Efficiency $\hat\theta$ $\times 100$ while output $y$ is top-1 accuracy, and inputs are $x_1$ is NVIDIA V100 latency, $x_2$  is ARM~Kirin~970 latency and $x_3$ is training cost.
    }
    \label{tab:rankdeter}
    \centering
    \footnotesize
    \setlength\tabcolsep{3pt} 
    
    \resizebox{\columnwidth}{!}{
    \begin{tabular}{l c c c c c c c c c c c c c c c c c} \\
        &  \rotate{\textbf{MobileNetV3}} & \rotate{\textbf{MobileNetV2}} & \rotate{\textbf{MobileNetV1}} & \rotate{\textbf{ShuffleNetV2}} & \rotate{\textbf{ShuffleNetV1}} & \rotate{\textbf{ResNet50}} & \rotate{\textbf{ResNet34}} & \rotate{\textbf{ResNet18}} & \rotate{\textbf{GoogleNet}} & \rotate{\textbf{SqueezeNetV1}.0~~} & \rotate{\textbf{SqueezeNetV1.1}~~} & \rotate{\textbf{InceptionV3}} & \rotate{\textbf{ResNeXt50}} & \rotate{\textbf{MNASNetV1}} & \rotate{\textbf{DenseNet121}} & \rotate{\textbf{DualPathNet92}} & \rotate{\textbf{VGG11}}  \\
     \hline
     { $x_1, x_2$}& 100 & 69 & 73 & 87 & 80 & 35 & 70 & 85 & 37 & 66 & 100 & 25 & 20 & 68 & 17 & 10 & 21 \\
     { $x_1,x_2,x_3$} & 100 & 69 & 79 & 100 & 88 & 37 & 70 & 86 & 50 & 66 & 100 & 26 & 24 & 68 & 25 & 12 & 27\\ 
\end{tabular}
}
\end{table}

\begin{figure*}[!t]
    \centering
    \includegraphics[width=0.55\textwidth]{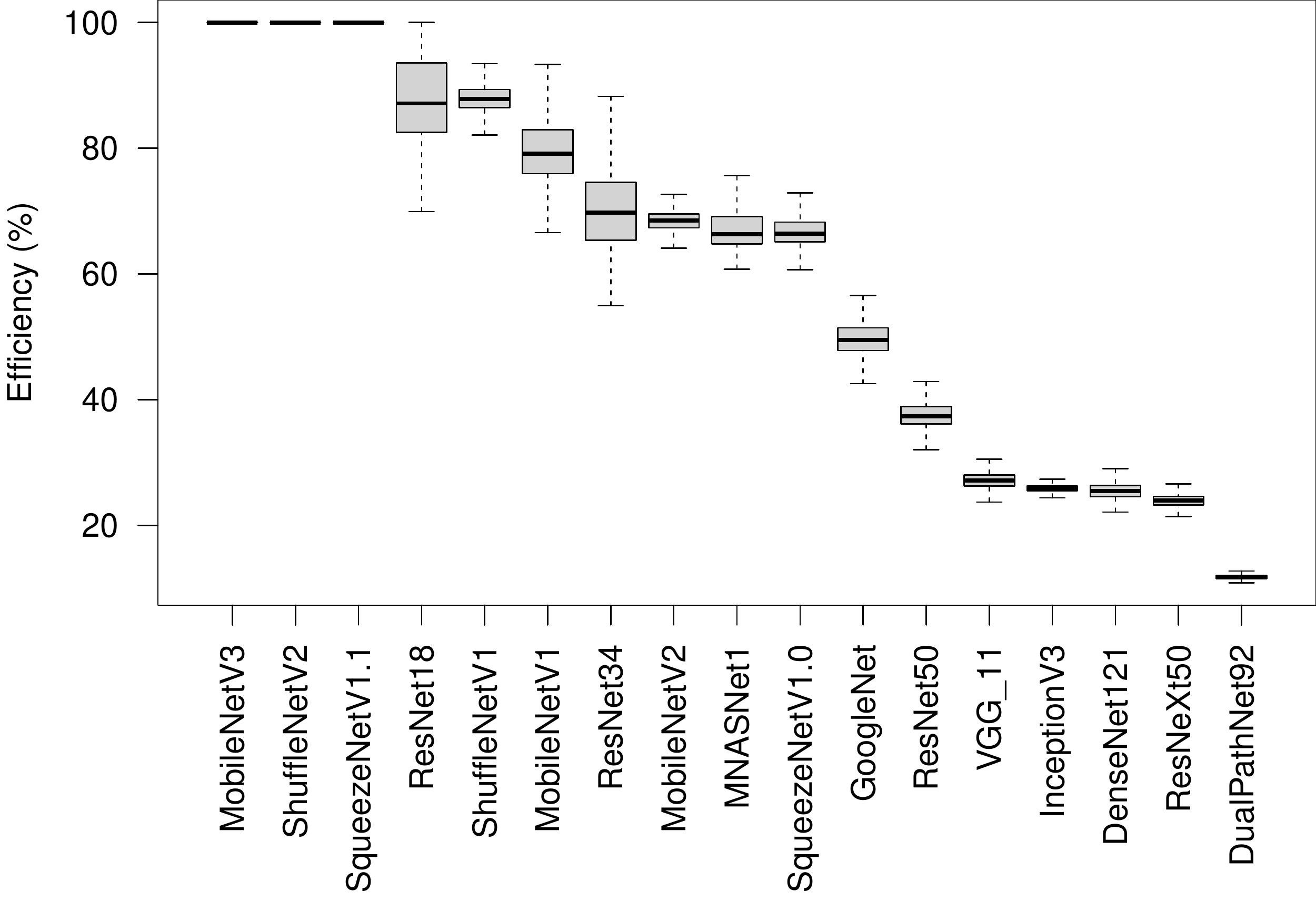}~~~~~~
    \includegraphics[width=0.39\textwidth]{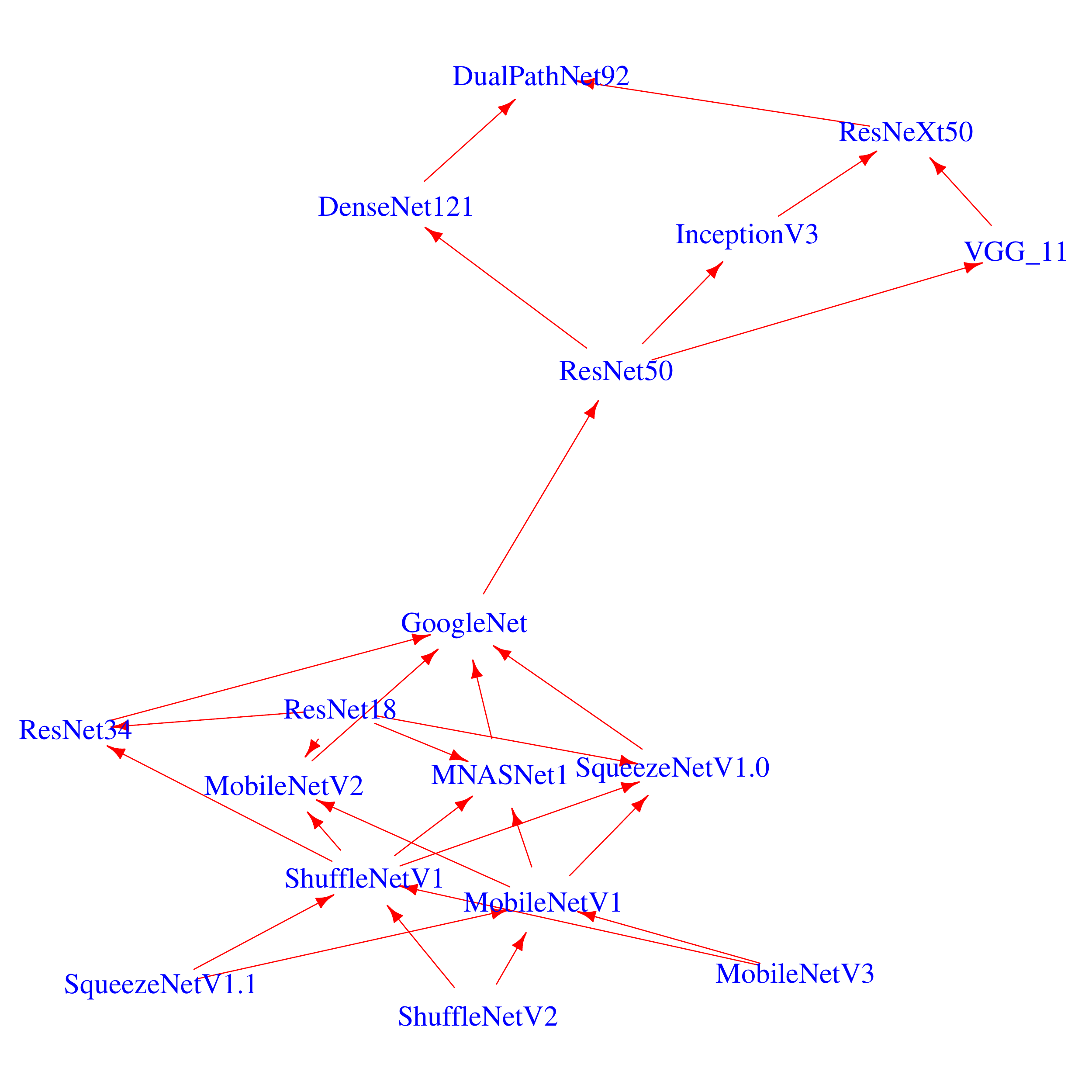}\\
    \small{(a)}\hspace{7cm}\small{(b)}\\
    
    \caption{ (a) Estimated random efficiency ($\hat\theta$) boxplots  ($y$ : top-1 accuracy, $x_1$: NVIDIA V100 latency, $x_2$: ARM~Kirin~970 latency, $x_3$: Training cost). (b) The stochastic dominance diagram for various image classification models}
    \label{fig:randtheta}
\end{figure*}

To rank deep learning models in terms of hardware efficiency, one needs to include as many computation complexity parameters as possible to reflect different  hardware targets. 
Here, we consider training cost and inference latency on different hardware targets as the main objectives of our optimization.
\textit{Inference latency} on a specific hardware is often of interest and is a good reflection of the actual performance on that specific hardware. In Table~\ref{tab:rankdeter} we propose to measure the inference latency directly on cloud (NVIDIA V100) and edge hardware (ARM Kirin 970) as the inputs, while keeping training cost on ImageNet1K dataset as the common \textit{training} complexity measure.  

Surely, a deep learning model requiring more training cost to obtain a given classification score is less desirable. Note that the \textit{training cost} is largely ignored by the community as a measure of model complexity. Figure~\ref{fig:unipareto} emphasizes the fact that training cost can change the behaviour of the Pareto frontier dramatically.    In this Figure, the efficiency varies depending on the CPU latency (Figure~\ref{fig:unipareto}(a)), or training cost (Figure~\ref{fig:unipareto}(b)) as \textit{univariate} inputs. The most efficient  network is MobileNetV3 and MNASNetV1 if  CPU latency is the single input, but changes to ShuffleNetV2 if training cost is the single input. Although, Figure~\ref{fig:unipareto} illustrates the univariate Pareto frontier, Table~\ref{tab:rankdeter} shows the efficiency as a multi-dimensional case, where latency on the edge, latency on the cloud as well as training cost are considered as the inputs ($\x_i$) and accuracy is the single output ($y_i$).

A close look at Table~\ref{tab:rankdeter} reveals that several models end up with a close efficiency. Manually-designed tiny models such as MobileNetV3 and SqueezeNetV1.1 are efficient when only inference is considered. However, MobileNetV1 becomes efficient when training cost is added to the inputs confirming that it is easier to train.

The deterministic efficiencies reported in  Table~\ref{tab:rankdeter} are calculated while assuming multi-dimensional inputs and top-1 accuracy as the single output. However, in practice, the model accuracy, latency, and training cost are random  and vary at each deployment. In order to take into account this randomness, we repeated the experiment $k=5$ times and then used the parametric bootstrap method  (i.e. Algorithm~\ref{alg:bootstrap}) to estimate random efficiency distributions. The boxplots in Figure~\ref{fig:randtheta}(a) shows the relative efficiency distributions, where $y$ is top-1 accuracy, $x_1$ is NVIDIA V100 latency, $x_2$ is ARM~Kirin~970 latency, and $x_3$ is training cost. Smaller networks such as ShuffleNetV2, MobileNetV3, and SqueezNetV1 are dominant, since their boxplot whiskers are very narrow and close to 100\%. Some average size networks such as Resnet18 and MobileNetV2 produce very similar efficiencies  since their boxplots overlap. Finally, DualPathNet92 exhibits the worse performance as its boxplot is below the 20\%  efficiency threshold. Figure~\ref{fig:randtheta}(b) summarize the stochastic dominance of the deep learning models used in this experiment. In this diagram, the model at the beginning of the arrow dominates the model at the end of the arrow. This means the lower constellation of the models always dominate the upper constellation if we include $x_1, x_2$, and $x_3$ as inputs in the measures of efficiency on any hardware platform.

\section{Conclusion}
We proposed a linear optimization framework to measure the relative efficiency of various deep learning models using empirical measurements of various competing performance metrics such as accuracy, latency, and training cost. The multi-dimensional Pareto frontier, as a by-product of redefining the relative efficiency, now accounts for different objectives such as cost of training, latency, and also their random variations. Our proposed method distinguishes deep learning models that run efficiently on a wide range of computing hardware. Moreover, it incorporates low training cost, while accounting for efficient inference, and more importantly while considering their random fluctuations.

\nocite{langley00}

\bibliography{main}
\bibliographystyle{icml2022}

\newpage


\end{document}